\pdfoutput=1

\documentclass[11pt]{article}

\usepackage[final]{acl}

\usepackage{times}
\usepackage{latexsym}
\usepackage{tcolorbox}
\usepackage{colortbl}
\usepackage{xcolor}

\usepackage[T1]{fontenc}

\usepackage[utf8]{inputenc}

\usepackage{microtype}

\usepackage{inconsolata}

\usepackage{graphicx}
\usepackage{subfig}
\usepackage{amsmath}
\usepackage{amssymb}
\usepackage[utf8]{inputenc} 
\usepackage{CJK}
\usepackage{booktabs}
\usepackage{xcolor}
\usepackage{enumitem}

\title{Instance-level Randomization: Toward More Stable LLM Evaluations}

\author{
    \textbf{Yiyang Li\textsuperscript{1,}}\thanks{\; Corresponding author.}, 
    \textbf{Yonghuang Wu\textsuperscript{2}}, 
    \textbf{Ying Luo\textsuperscript{1}}, 
    \textbf{Liangtai Sun\textsuperscript{1}}, 
\\
    \textbf{Zishu Qin\textsuperscript{2}}, 
    \textbf{Lin Qiu\textsuperscript{1}}, 
    \textbf{Xuezhi Cao\textsuperscript{1}}, 
    \textbf{Xunliang Cai\textsuperscript{1}}
\\
\\
    \textsuperscript{1}Meituan Group, 
    \textsuperscript{2}Fudan University 
\\
\small{
 \{liyiyang06,luoying08,sunliangtai,qiulin07,caoxuezhi,caixunliang\}@meituan.com}
 \\
\small{
 \{yonghuangwu21,zsqin23\}@m.fudan.edu.cn}
}

\begin{document}
\begin{CJK}{UTF8}{gbsn}

\maketitle
\begin{abstract}
Evaluations of large language models (LLMs) suffer from instability, where small changes of random factors such as few-shot examples can lead to drastic fluctuations of scores and even model rankings. Moreover, different LLMs can have different preferences for a certain setting of random factors. As a result, using a fixed setting of random factors, which is often adopted as the paradigm of current evaluations, can lead to potential unfair comparisons between LLMs. To mitigate the volatility of evaluations, we first theoretically analyze the sources of variance induced by changes in random factors. Targeting these specific sources, we then propose the instance-level randomization (ILR) method to reduce variance and enhance fairness in model comparisons. Instead of using a fixed setting across the whole benchmark in a single experiment, we randomize all factors that affect evaluation scores for every single instance, run multiple experiments and report the averaged score. Theoretical analyses and empirical results demonstrate that ILR can reduce the variance and unfair comparisons caused by random factors, as well as achieve similar robustness level with less than half computational cost compared with previous methods. Codes and data are available at \url{https://github.com/EricLee8/Instance-level-Randomization}.
\end{abstract}

\section{Introduction}
\begin{figure*}[tbp]
    \centering
    \includegraphics[width=0.92\textwidth]{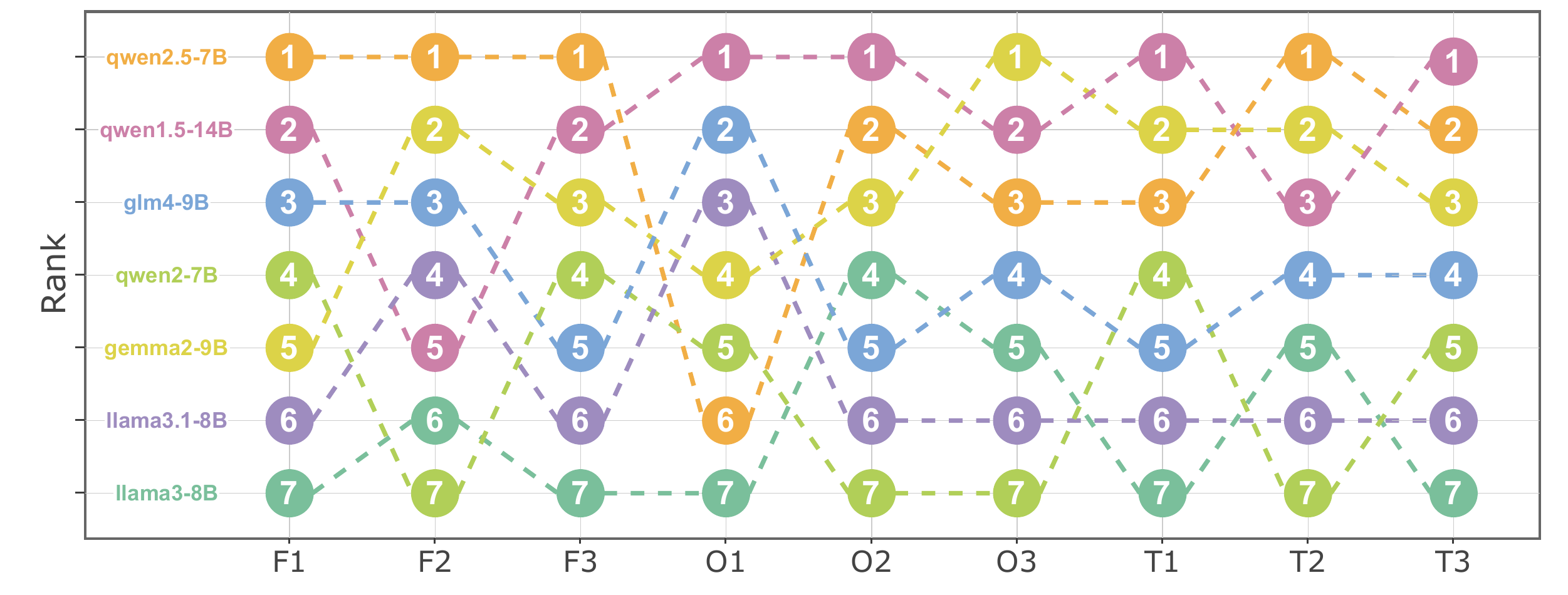}
    \caption{Model ranking variation on Hellaswag with different random factors changing, where ``F'' represents few-shot examples, ``O'' represents option labels, and ``T'' represents task descriptions.}
    \label{fig:model_ranking_vary}
\end{figure*}

With large language models (LLMs) getting stronger~\cite{gpt4o_system_card,deepseek3_technical_report,qwen3_technical_report,the_llama4_herd}, evaluation of them is also becoming harder and more important~\cite{a_systematic_survey_and_critical_review, a_survey_on_evaluation}. One of the biggest challenges of evaluation is that it suffers from instability, where small changes of random factors such as few-shot examples, task descriptions, and even evaluation frameworks can result in significant fluctuations of evaluation scores, and even model rankings~\cite{qian-etal-2024-varbench,does_fewshot_learning_help_llm,the_order_effect,when_benchmarks_talk}. Figure\ref{fig:model_ranking_vary} illustrates an example where $7$ LLMs are evaluated on Hellaswag~\cite{can_a_machine_really_finish_your_sentence} dataset. As random factors vary across different runs, model rankings change drastically, with Qwen2.5-7B ranking over as large a range as from 1st to 6th.

Given the above observations, literature can be categorized into the following two groups. One group of works mainly focus on reporting the extreme values or range of fluctuations caused by these random factors. They either utilize Automated Prompt Engineering to find the best or worst prompts, and report the corresponding scores under these prompts~\cite{on_the_worst_prompt_performance}, or design methods to estimate the range of evaluation scores~\cite{quantifying_variance,efficient_multiprompt_evaluation,when_benchmarks_are_targets}. Another group of works pay attention on mitigating the instability before the evaluation phase. For example, \citealt{improving_prompt_consistency} adopt constrained decoding to make outputs more consistent, \citealt{meaning_typed_prompting} suggest formatting the multiple-choice tasks as cloze tests to avoid the unstable decoding process.

However, few works attempt to enhance the robustness of evaluation~\cite{towards_llms_robustness}. In this paper, we fill the gap of stabilizing evaluations by proposing an embarrassingly simple yet surprisingly effective method to mitigate the bias and variance of evaluations caused by random factors.

In terms of the most adopted paradigm of the current evaluations, it usually adopts a fixed setting of random factors and only runs a single experiment, which imposes biased settings, high variance, and potential unfair comparisons between models due to LLMs' volatility on the random factors. A natural and naive alleviation is to vary the settings of random factors across multiple runs and report the averaged results~\cite{a_call_for_multiprompt_llm_evaluation}. However, this method suffers a high variance and requires a large number of runs to obtain an unbiased and stable result. Based on this, we theoretically analyze the sources of the evaluation instability, based on which we propose our instance-level randomization (ILR) method to stabilize evaluations. Specifically, instead of using a fixed setting for a single experiment, we apply random settings to each of the instances in a benchmark. We show both theoretically and empirically that our proposed ILR is able to reduce the variance faster, using less than $50\%$ computational cost to obtain the same variance.

In this paper, we study few-shot examples, task descriptions, prompt formats, and option labels as the random factors that affect evaluations, whose detailed explanations are in Appendix \ref{appendix:diff_random_factors}. Nevertheless, we underline that without the loss of generality, our ILR can be applied to as many as random factors as we can consider.

To sum up, the contributions of this work are:
\begin{itemize}[itemsep=0pt, parsep=0pt, topsep=0pt, partopsep=0pt]
    \item We propose comprehensive theoretical analyses of the sources of the evaluation instability caused by the random factors.
    \item Based on the theoretical analyses, we propose a simple yet effective instance-level randomization (ILR) method to obtain a more robust evaluation, which can achieve the same robustness level with less than half computational cost compared with previous methods.
    \item Empirical results solidly validate the theoretical analyses, as well as the effectiveness of our method.
\end{itemize}

\section{Related Work}

Instability of evaluations have been discovered and studied by the community for a long time. Studies have shown that random factors such as phrasing and wording of prompts~\cite{what_did_i_do_wrong, benchmarking_prompt_sensitivity}, choice and order of few-shot examples~\cite{does_fewshot_learning_help_llm, pisarevskaya2025zero}, as well as their format and structure~\cite{towards_llms_robustness}, can lead to large variances in model outputs~\cite{prompting_science_report_1}, even affecting factual accuracy~\cite{evaluating_llms_in_materials_science}. Given these factors, studies focus on either reporting or mitigating the evaluation instability. 

\subsection{Reporting Evaluation Instability}

With growing awareness of instability in LLM evaluations caused by random factors, researchers are reporting this phenomenon by quantifying and representing its extent. Studies explore performance under extreme conditions, identifying ``best'' prompts through Automated Prompt Engineering~\cite{hard_prompts_made_easy,prompt_engineering_a_prompt_engineer, prompt_rewriting_with_reinforcement_learning, a_survey_of_automatic_prompt_engineering} and ``worst'' prompts~\cite{on_the_worst_prompt_performance}. However, focusing on best or worst prompts may not reflect real-world experiences where prompts are diverse and spontaneous, and they require much additional compute, posing challenges for scalable and generalizable evaluation.

Additionally, compared to directly reporting extreme values or variances~\cite{accounting_for_variance, quantifying_variance, eavluating_llms_on_nonfunctional_requirements}, researchers have proposed other statistical metrics, such as performance gaps across different prompts~\cite{quantifying_llms_sensitivity, a_call_for_multiprompt_llm_evaluation}, or median performance and percentile-based metrics~\cite{efficient_multiprompt_evaluation}. These methods provide valuable insights into the statistical characteristics of the evaluation instability, but they only focus on the phenomenon itself, rather than offering direct strategies to alleviate its impact on the evaluation process.

\subsection{Mitigating Evaluation Instability}

Recognizing the presence of instability in LLM evaluations and its potential impact on model rankings, several approaches are proposed to mitigate instability by modifying model outputs. MTP~\cite{meaning_typed_prompting} enhances the reliability and consistency of structured output generation by constrained decoding, but it primarily focuses on stabilizing the generation process rather than the evaluation itself. Similarly, \citealt{quantifying_variance} propose to reframe multiple-choice tasks as cloze filling to avoid the unstable generation process, thus reducing variance.

Most related to our work, MOF~\cite{towards_llms_robustness} enhances robustness to prompt format changes by diversifying expression styles in few-shot examples. While effective in reducing performance gaps caused by formatting variations, MOF focuses solely on non-semantic aspects of few-shot examples. In contrast, our approach is applicable to a broader range of random factors, offering a more comprehensive solution to mitigating the evaluation instability.

In summary, existing methods lack the generality to tackle diverse instability sources. Our work fills this gap by proposing a unified framework that stabilizes evaluations against a wide array of random factors.

\section{Problems of Current Evaluation}
\label{sec:problems_of_cur_eval}

\begin{figure}[tbp]
    \includegraphics[width=0.49\textwidth]{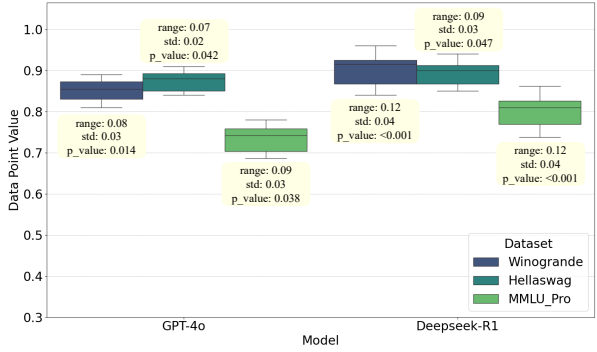}
    \centering
    \caption{Box plots of different LLMs' scores on the Winogrande, Hellaswag, and MMLU-Pro datasets, where each model is evaluated $8$ times under various few-shot example settings. The box plots illustrate the distribution of scores across these $8$ runs, where the yellow text box around each box-plot represents the statistical analysis on the results.} 
    \label{fig:box_plot}
\end{figure}

In this section, we show that current evaluation paradigm of LLMs suffers from biased settings, high variance, and potential unfair comparisons between different models. Specifically, there are many random factors that can affect the performance scores of LLMs in a benchmark, such as instructions, task descriptions, prompt formats, and few-shot examples. However, current benchmarks usually adopt a fixed setting of the above random factors, resulting in biased and unstable evaluations of different LLMs.

\subsection{Biased Settings}
In real-world scenarios, different users may use different settings of the aforementioned random factors to prompt LLMs, which leads to a discrepancy between benchmark scores with fixed factors and the actual user experience.

Formally, the true average score of LLMs on a benchmark should be calculated as follows:
\begin{equation}
\label{eq:ideal_eval}
    S_M = \mathbb{E}_{f \sim \mathcal{F}} \; \mathbb{E}_{(x, y) \sim D} \; \mathbb{I}(M(f(x)) = y),
\end{equation}
where $S_M$ means the score of model $M$, $f$ is sampled from $\mathcal{F}$, which is the joint distribution of all random factors that may affect the benchmark performance, $(x, y)$ is the query, where problem $x$ and reference answer $y$ are sampled from the dataset distribution $D$, and $\mathbb{I}$ is an indicator function that outputs $1$ if the output of the LLM matches the reference $y$.

Current benchmarks only adopt a fixed setting of random factors $f_{biased}$, resulting in biased estimation of the true average score:
\begin{equation}
    S^{biased}_M = \mathbb{E}_{(x, y) \sim D} \; \mathbb{I}(M(f_{biased}(x)) = y).
\end{equation}
Different choice of $f_{biased}$ can lead to significant changes of benchmark scores, and even shift the model rankings.

\subsection{High Variance}
Using a fixed setting to evaluate LLMs on a benchmark also introduces high variances of evaluation scores. We show this observation with an example of few-shot evaluations on two strongest models at present: GPT-4o~\cite{gpt4o_system_card} and Deepseek-R1~\cite{deepseek-r1}. Specifically, we conducted $8$ experiments for each model on the Winogrande~\cite{an_adversarial_winograd_schema_challenge}, Hellaswag~\cite{can_a_machine_really_finish_your_sentence}, and MMLU-Pro~\cite{mmlu-pro} datasets using different few-shot examples, with each dataset containing $100$ instances in Winogrande and Hellaswag, and $500$ instances in MMLU-Pro.

Figure \ref{fig:box_plot} presents box plots of the scores achieved by various LLMs on the three datasets, where the p-value shown in the yellow text box around each box-plot represents the statistical significance of the difference between the best and worst runs out of the $8$ experiments, as determined by a paired t-test. As shown in the figure, the choice of few-shot examples has a substantial impact on evaluation scores across all datasets even with the strongest models: the maximum range between the best and worst runs can reach $12\%$, and in all cases, the differences between the highest and lowest scores are statistically significant ($p < 0.05$).

This observation demonstrates that results obtained from a single experiment with a fixed setting can have high variance, making score comparisons between models unreliable. For more examples of other random factors and on other models, please refer to Appendix \ref{appendix:box_plot}.

\begin{figure}[tbp]
    \centering
    \subfloat{\includegraphics[width=0.40\textwidth]{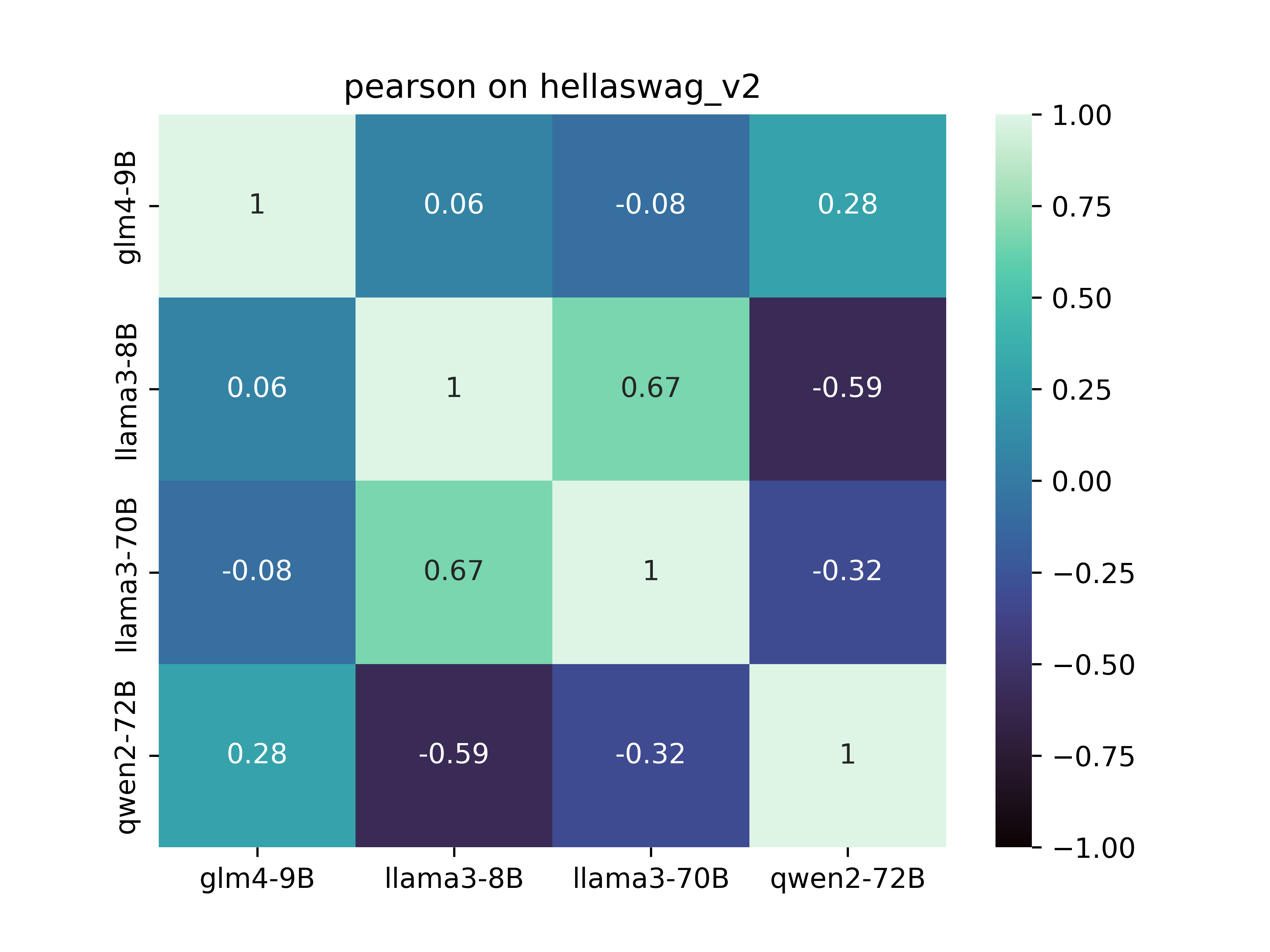}}%
    \hspace{0.01\textwidth}%
    \subfloat{\includegraphics[width=0.40\textwidth]{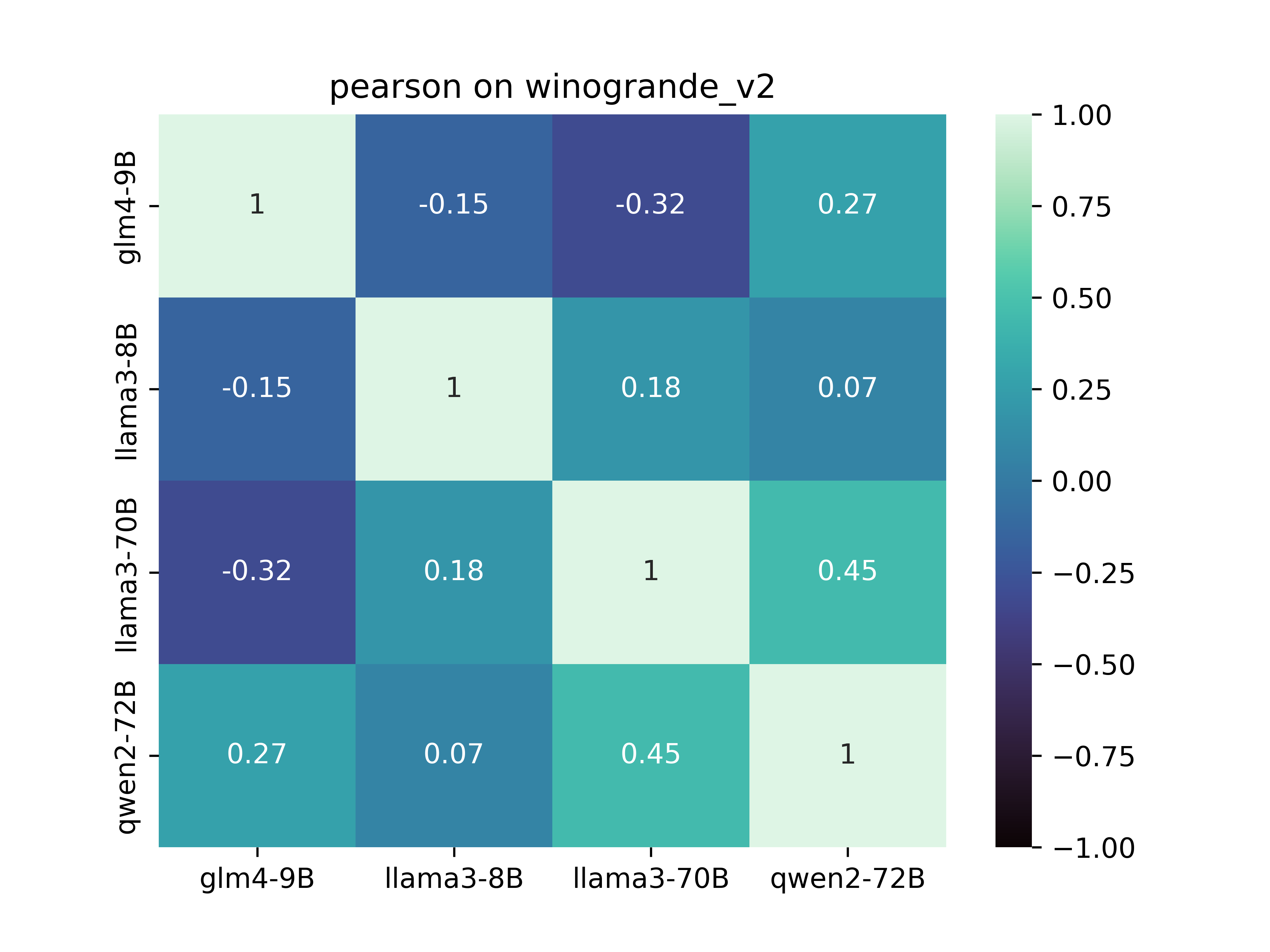}}
    \caption{Correlations of different LLMs over different few-shot examples on Hellaswag and Winogrande.}
    \label{fig:corr}
\end{figure}

\subsection{Unfair Comparisons}
Another problem of the current fix-setting evaluation is potential unfair comparisons between LLMs. Specifically, different models may exhibit distinct preferences for particular random factors. Taking few-shot evaluation as an example, if we use a fixed set of few-shot examples to evaluate Model A and Model B, we may observe that Model A outperforms Model B. However, if we switch to another set of few-shot examples, the result may be reversed. More generally, this observation manifests as low or even negative correlations between models when multiple experiments are conducted with different random factors.

Figure \ref{fig:corr} illustrates the Pearson correlations of different LLMs over different few-shot examples on Hellaswag and Winogrande. We can see from the figure that the correlations are relatively low, where Llama3-8B and Qwen2-72B even show a moderately strong negative correlation. This indicates that using a fixed setting to evaluate models can be potentially unfair if the fixed setting happens to be preferred by one model, and disfavored by another. For correlation analysis of more random factors, please refer to Appendix \ref{appendix:corr}.

\begin{figure*}[tbp]
    \centering
    \includegraphics[width=0.98\textwidth]{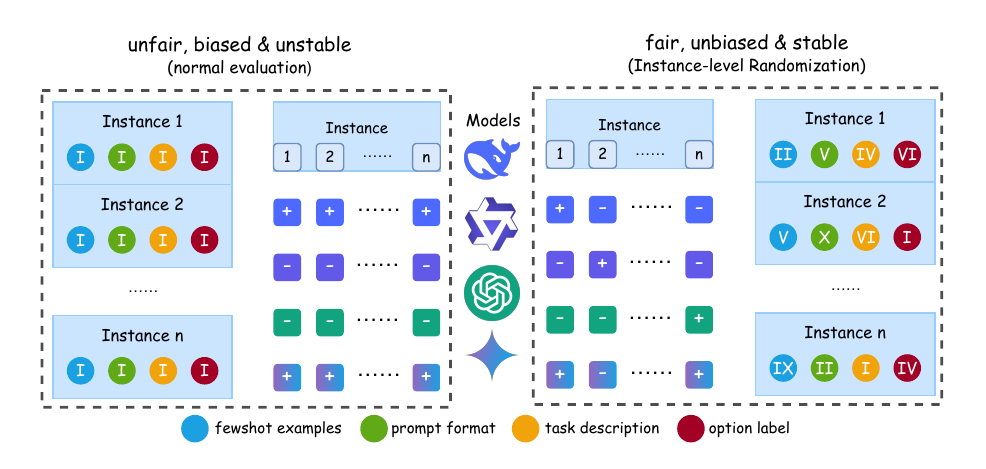}
    \caption{A brief illustration of our ILR method. Different Roman numerals in circles represent different evaluation settings applied to each instance of a benchmark. The plus and minus signs mean that a certain evaluation setting has positive or negative effect on the evaluation result of an instance for a specific model. Different colors of the signs represent different models, who have specific preferences over certain evaluation settings.}
    \label{fig:methodoloy}
\end{figure*}

\section{Methodology}
In this section, we propose a simple yet effective method using \textbf{I}nstance-\textbf{L}evel \textbf{R}andomization (ILR) over random factors, which can reduce the variance of evaluation scores, and mitigate the instability of model rankings at a lower cost. Figure \ref{fig:methodoloy} briefly illustrates our ILR method and the comparison between ILR and the normal evaluation setting. The left side illustrates the normal evaluation paradigm, in which the same set of random factor settings is applied to every instance within a benchmark. Since a given model tends to exhibit consistent preferences toward the same settings, each instance in the benchmark tends to simultaneously overestimate or underestimate the model's true accuracy for that instance. This leads to unfair, biased, and unstable evaluation results. In contrast, the right side presents our proposed ILR method, which employs different settings for all random factors for each instance. This approach offsets preference effects against each other, resulting in fairer, more stable, and less biased evaluation outcomes.

In the subsequent sections, we provide both theoretical and empirical evidence demonstrating that ILR achieves a faster reduction in evaluation variance with fewer runs, thus lowering computational overhead. Besides, we also introduce the Observed Reversal Probability (ORP) as a novel metric to quantify the stability of model rankings. Our experiments confirm that ILR significantly reduces ORP, proving its ability to produce fairer, more stable, and ultimately more reliable evaluation results.


\subsection{Variance Analysis}
\label{sec:variance_analysis}
Before introducing our ILR method, we analyze the sources of variance when using the mean of $n$ evaluation runs, each considering different random factors, as the benchmark evaluation result. We rewrite Eq. \eqref{eq:ideal_eval} from its expected value form to its sampled mean form. Suppose for the $i_{th}$ evaluation run, the setting of the random factors is $f_i$, the corresponding accuracy of the benchmark containing $m$ instances $\{(x_k, y_k)\}_{k=1}^m$ is marked as $\bar{A_i} = \frac{1}{m}\sum_{k=1}^m\mathbb{I}(M(f_i(x_k))=y_k)$, and the final average accuracy of the benchmark is $\bar{A} = \frac{1}{n}\sum_{i=1}^n\bar{A_i}$. For simplicity, we write $\mathbb{I}(M(f_i(x_k))=y_k)$ as $f_i(x_k)$. So far, we can calculate the variance of the mean of $n$ runs as follows:
\begin{equation}
\label{eq:variance_analysis}
\tiny
\begin{aligned}
    Var(\bar{A}) &= Var\left(\frac{1}{n}\sum_{i=1}^n \bar{A}_i \right) \\
    &= \frac{1}{n^2} \left( \sum_{i=1}^n Var(\bar{A}_i) + 2 \sum_{1 \leq i < j \leq n} Cov(\bar{A}_i, \bar{A}_j) \right),\\
    Var(\bar{A}_i) &= \frac{1}{m^2} \bigg( \sum_{k=1}^m Var(f_i(x_k)) \\
    &\hspace{2cm} + 2 \sum_{1 \leq k < l \leq m} Cov(f_i(x_k), f_i(x_l)) \bigg).
\end{aligned}
\end{equation}
Combining the equations in Eq. \eqref{eq:variance_analysis}, we have:
\begin{equation}
\label{eq:expanded_variance}
\tiny
\begin{aligned}
    Var(\bar{A}) = &\frac{1}{n^2}\frac{1}{m^2}\sum_{i=1}^n\sum_{k=1}^m Var(f_i(x_k))\ + \\
    & \frac{1}{n^2}\frac{1}{m^2}\sum_{i=1}^n\sum_{1\le k < l\le m}2\ Cov(f_i(x_k), f_i(x_l))\ + \\
    & \frac{1}{n^2}\sum_{1\le i < j\le n} 2\ Cov(\bar{A_i}, \bar{A_j}).
\end{aligned}
\end{equation}
Using $\bar{A_i} = \frac{1}{m}\sum_{k=1}^mf_i(x_k)$ and the linearity of covariance, $Cov(\bar{A_i}, \bar{A_j})$ can be expanded as follows:
\begin{equation}
\label{eq:dataset_cov_expanded}
\tiny
Cov(\bar{A_i}, \bar{A_j}) = \frac{1}{m^2}\sum_{k=1}^m\sum_{l=1}^m Cov(f_i(x_k), f_j(x_l)).
\end{equation}

There are three items that consist of $Var(\bar{A})$, each of which is one line in Eq. \eqref{eq:expanded_variance}. The first item is the average of the inherent variance of each instance's accuracy, which follows a Bernoulli distribution and cannot be optimized since it only depends on its mean $p$. The second term represents the pairwise covariance between instances within the dataset, which is typically positive, as instances from the same dataset are usually positively correlated. Reducing the correlation between instances can help minimize this term. The third term is the pairwise covariance of experiment scores across runs with different random factors. Similarly, reducing the correlation between experiments can help minimize this term.

\subsection{Instance-level Randomization}
\label{sec:ilr}
To start with, let us revisit the meaning of Eq. \eqref{eq:ideal_eval} and the current evaluation convention first. We sample a setting of random factors $f_{biased}$ from the combinational distribution $\mathcal{F}$, fix it across the sampled data points $(x, y)$ from the data distribution $D$, and calculate the mean accuracy of these data points as the final score. This convention leads to the aforementioned three problems in Section \ref{sec:problems_of_cur_eval}.

To mitigate the problems causing by a fixed $f_{biased}$, previous works propose to run multiple experiments of different settings and report the average scores~\cite{a_call_for_multiprompt_llm_evaluation}, or report confidence intervals and range distributions~\cite{quantifying_llms_sensitivity, efficient_multiprompt_evaluation}. However, they require relatively high number of experiments or more compute to reduce the variance to an acceptable range. Different from previous works, this paper aims to identify methods that can reduce evaluation variance more rapidly with fewer runs.

As mentioned in Section \ref{sec:variance_analysis}, the most effective way to reduce variance is to decrease the correlation between instances within the benchmark and the correlation between multiple runs of the benchmark. Therefore, we summarize two points to ensure unbiased and fair evaluation, as well as reduce variance:
\begin{itemize}[itemsep=0pt, parsep=0pt, topsep=0pt, partopsep=0pt]
    \item We consider as much random factors as possible when sampling settings, which helps obtain a more unbiased performance estimation.
    \item Instead of using a fixed setting of random factors across the benchmark, we randomly sample different settings for each instance, which helps decrease the instance-wise correlation thus reducing variance, as well as mitigate potential unfair comparisons caused by different LLMs' preferences over a fixed setting of specific random factors.
\end{itemize}

Instance-level randomization changes the instance-wise covariance term in Eq. \eqref{eq:expanded_variance} to $Cov(f^k_i(x_k), f^l_i(x_l))$, which means different instances are now evaluated under different random factor settings within an experiment. Also, it changes the inner part of the experiment-wise covariance term in Eq. \eqref{eq:dataset_cov_expanded} to $Cov(f^k_i(x_k), f^l_j(x_l))$, which means the differences between experiments now arise not only from a single setting of random factors, but also from distinct random factors applied to each instance. This helps decrease the correlation between instances and also experiments. To explain intuitively, two instances evaluated with different random factors tend to be less correlated than those evaluated with the same random factor.

As for the variance term, the inner part of summation changes from $Var(f_i(x_k))$ to $Var(f_i^k(x_k))$. The change of random factors of an instance may lead to the change of its inherent probability of being answered correctly, thus changing the variance\footnote{The variance of Bernoulli distribution is $p(1-p)$.}. However, since the change of random factors is randomly sampled, we can assume that the average impact of the variance term to be $0$.\\

All of the above theoretical analyses are empirically validated in Section \ref{sec:cov_reduction}.

\subsection{Speed of Variance Reduction}
\label{sec:theoretical_var_speed}
In this section, we derive that considering more random factors can accelerate the rate at which the variance of $n$ experiments decreases as $n$ increases, while the ILR method can speed up the rate at which the variance of a single experiment decreases with regard to the number of dataset instances $m$. Although it is challenging to analyze theoretically how the variance decreases with $n$ when both approaches are combined, our empirical results in Section \ref{sec:faster_var_reduction} demonstrate that using both methods together can further accelerate the reduction of variance.

We can rewrite the first equation of Eq. \eqref{eq:variance_analysis} as its mean form:
\begin{equation}
\tiny
\begin{aligned}
    \overline{Var} &= \frac{1}{n}\sum_{i=1}^n Var(\bar{A_i}),\\
    \overline{Cov} &= \frac{2}{n(n-1)}\sum_{1\le i<j \le n} Cov(\bar{A_i}, \bar{A_j}),\\
    Var(\bar{A}) &= \frac{1}{n^2} \left(n \overline{Var} + 2\frac{n(n-1)}{2}\overline{Cov} \right)\\
    & = \frac{1}{n}\overline{Var} + \frac{n-1}{n}\overline{Cov}\\
    & = \frac{1}{n}(\overline{Var} - \overline{Cov}) + \overline{Cov}.
\end{aligned}
\end{equation}
When considering more random factors, the average variation of single experiment $\overline{Var}$ is unchanged, while the average covariance $\overline{Cov}$ between experiments is reduced. The derivative of $Var(\bar{A})$ with regard to $n$ is $-\frac{1}{n^2}(\overline{Var} - \overline{Cov})$, the absolute value of which will be increased with the decrement of $\overline{Cov}$ since usually we have $\overline{Var} > |\overline{Cov}|$. Therefore, we can achieve lower variance with smaller $n$. Similarly, we can derive that the speed of single experiment variance reduction with regard to $m$.

\begin{table*}
\centering
\begin{tabular}{c|c c c c c c}
    \specialrule{0.09em}{0.0pt}{0.3pt}
     & $\text{Corr}_{\text{instance}}^{\text{fixed}}$ & $\text{Corr}_{\text{instance}}^{\text{random}}$ & $\text{Corr}_{\text{experiment}}^{\text{fixed}}$ & $\text{Corr}_{\text{experiment}}^{\text{random}}$ & $\text{Var}_{\text{instance}}^{\text{fixed}}$ & $\text{Var}_{\text{instance}}^{\text{random}}$ \\ \hline
    Winogrande & 0.177 & 0.091 & 0.096 & 0.073 & 0.207 & 0.198 \\
    Hellaswag & 0.457 & 0.119 & 0.213 & 0.161 & 0.204 & 0.197 \\
    \specialrule{0.09em}{0.3pt}{0.0pt}
\end{tabular}
\caption{Correlation and variance in Eq. \eqref{eq:expanded_variance} before (marked as \textit{fixed}) and after (marked as \textit{random}) instance-level randomization.}
\label{tab:cov_reduction}
\end{table*}

\section{Experiments}
\label{sec:exp}
In this section, we first conduct experiments to empirically validate the theoretical analyses in Section \ref{sec:variance_analysis} and \ref{sec:theoretical_var_speed}. After that, we propose a metric called Observational Reversal Probability (ORP) to measure the effectiveness of ILR. The experiments are conducted on Winogrande, Hellaswag, BigBench-Hard~\cite{challenging_bigbench_tasks}, and MMLU-pro~\cite{mmlu-pro}, with $100$ samples on each subset of these datasets if applicable.

\subsection{Covariance Reduction}
\label{sec:cov_reduction}
To validate the theoretical analysis of covariance reduction in Section \ref{sec:ilr}, we conduct experiments under fixed settings and ILR settings, and calculate the correlation coefficients at both instance-level and experiment-level, which correspond to the second line of Eq. \eqref{eq:expanded_variance}, and Eq. \eqref{eq:dataset_cov_expanded}, respectively. Specifically, we vary a random factor as the cause of the variance, and fix or randomize others for fixed and ILR settings.

As shown in Table \ref{tab:cov_reduction}, we observe reduction of correlation at both instance-level and experiment-level, where the former drops more significantly after ILR. This is because prior to randomization, instance-level correlations mainly arise from shared random factors; after randomization, these correlations decrease more substantially. In contrast, experiment-level random factors are already different across experiments, even though they are fixed within each experiment, so their correlations are inherently lower and decrease less noticeably after randomization compared to the instance-level. As for the variance term of each instance itself, it remains relatively stable as analyzed in Section \ref{sec:ilr}.

\subsection{Faster Variance Reduction}
\label{sec:faster_var_reduction}
To empirically validate the theoretical analysis of faster variance reduction of the proposed method, we conduct experiments by calculating the mean's variance of $n$ experiments with regard to the increase of $n$. Specifically, we also vary a random factor as the cause of the variance of a single experiment, and adopt three scenarios according to the setting of other random factors in $n$ experiments:
\begin{itemize}[itemsep=0pt, parsep=0pt, topsep=0pt, partopsep=0pt]
    \item For each of the $n$ experiments, only one random factor is considered, which is the same setting in previous works~\cite{a_call_for_multiprompt_llm_evaluation, towards_llms_robustness}.
    \item For each of the $n$ experiments, multiple random factors are considered.
    \item We combine the instance-level randomization and multiple random factors consideration.
\end{itemize}
We conduct $20$ experiments, each of which is repeated $15$ times to calculate the standard deviation of a single experiment. For each $n$, we randomly select $n$ out of the $20$ experiments ($n \leq 15$) as the results of $n$ experiments, and repeat this selection process $30$ times to get a more accurate estimation. The mean of these selections is used as the standard deviation of the mean across multiple experiments.

\begin{figure}[tbp]
    \centering
    \includegraphics[width=0.48\textwidth]{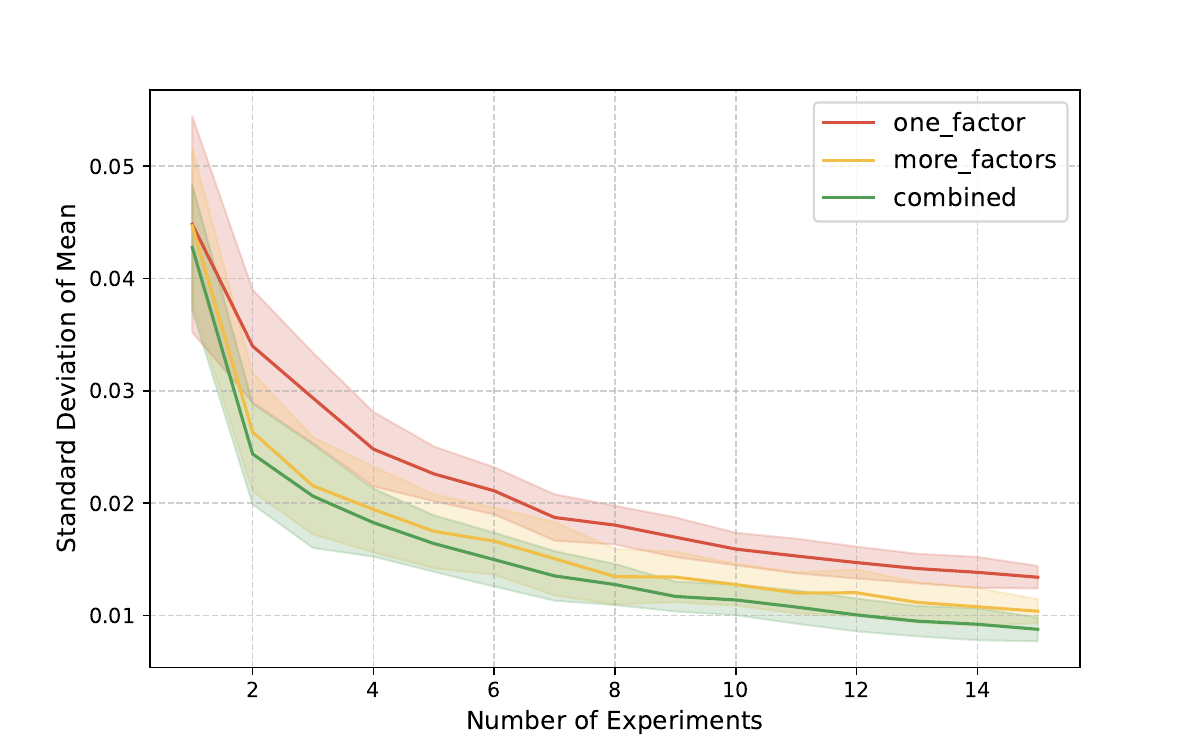}
    \caption{Variation reduction with the number of experiments. The solid lines represent the standard deviation of the mean over $n$ experiments, while the shaded area indicates the estimated standard deviation of this standard deviation, which is calculated from $30$ selections.}
    \label{fig:var_reduce_with_n}
\end{figure}

As shown in Figure \ref{fig:var_reduce_with_n}, the standard deviation drops as the number of experiments increases, which is in accordance with the Law of Large Numbers. Considering more random factors makes the standard deviation drop faster since it reduces the correlation between experiments, as theoretically analyzed in Section \ref{sec:theoretical_var_speed}. This observation is also in line with \citealt{accounting_for_variance}, where they study the effect of random factors on training. Additionally, applying ILR further accelerates the speed of standard deviation reduction. More intuitively, if we aim to reduce the standard deviation to $0.02$, previous methods require approximately $6.5$ experiments, whereas our ILR only needs about $3$. This reduces the evaluation cost by more than $\mathbf{50\%}$.

\subsection{Observed Reversal Probability}
Besides variance reduction, we show that ILR also helps obtain a more stable and fairer evaluation of model rankings, as well as achieve a more accurate estimation of the models' true performance on a benchmark.

To quantitatively analyze the impact of random factors on model rankings, we introduce the Observed Reversal Probability (ORP). ORP measures the likelihood that the observed ranking of two models (designated A and B) reverses due to the change of random factors.

Let the true accuracies of model A and B be $S_A$ and $S_B$, respectively. In practice, their observed accuracies are $O_A = S_A + \Delta S_A$ and $O_B = S_B + \Delta S_B$. Here $\Delta S_A$ and $\Delta S_B$ are normally distributed\footnote{Values affected by multiple random factors are usually normally distributed, e.g., scores of exams.} random variables with standard deviations $\sigma_A$ and $\sigma_B$, and a Pearson correlation coefficient $\rho_{AB}$. The observed performance difference, $\Delta_S = O_A - O_B$, consequently follows a normal distribution with a mean of $\delta_{AB} = S_A - S_B$ and a variance of $\sigma_A^2 + \sigma_B^2 - 2 \rho_{AB} \sigma_A \sigma_B$. ORP measures the probability that the observed ranking is reversed compared with the true ranking due to the change of random factors:
\begin{equation}
\label{eq:ORP}
\small
    \begin{aligned}
        ORP(\delta_{AB}) &= P(\delta_{AB}\Delta_S<0) \\
        &= \Phi\left(\frac{|\delta_{AB}|}{\sqrt{\sigma_A^2+\sigma_B^2-2 \rho_{AB} \sigma_A \sigma_B}}\right).
    \end{aligned}
\end{equation}
Here, $\delta_{AB}$ denotes the true difference between the two models, while $\Delta_S$ represents the observed score difference. A product of them less than zero indicates that the observed ranking is reversed compared to the true ranking. Therefore, the lower ORP, the more stable an evaluation. In practice, $\sigma_A^2$, $\sigma_B^2$, and $\rho_{AB}$ can be calculated from multiple runs. However, the true difference $\delta_{AB}$ is unknown. Therefore, we consider ORP as a function of $\delta_{AB}$ and calculate the area under the $ORP-\delta_{AB}$ curve (AUC) as the final ORP value between two models.

\begin{figure*}[tbp]
    \centering
    \includegraphics[width=0.97\textwidth]{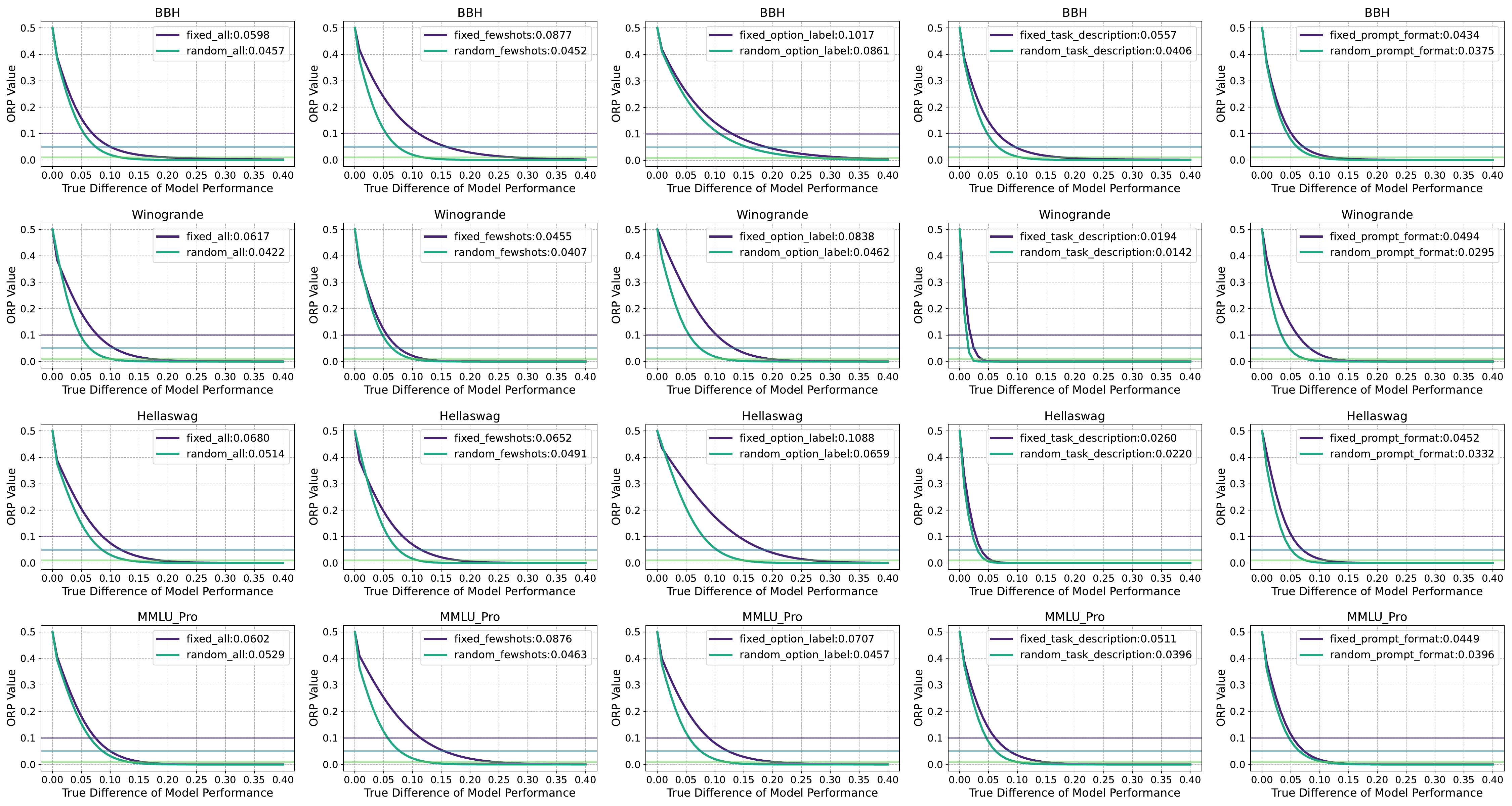}
    \caption{ORP curves between different LLMs on Hellaswag, Winogrande, BBH, and MMLU-Pro. We consider four random factors: few-shot examples, option labels, task descriptions, and prompt formats. Three horizontal lines indicate how much the difference of scores should be to have $90\%$, $95\%$, and $99\%$ confidence that one model actually outperforms another, respectively. The values in each legend are the AUCs. The lower the curves (smaller AUCs) the better.}
    \label{fig:orp_curve}
\end{figure*}

\subsection{ORP Reduction}
To show ILR makes the evaluation more stable, we conduct experiments by varying random factors and calculate the average ORP value between each pair of four models\footnote{Llma3-8B, GLM4-9B, Llama3-70B, and Qwen2-72B.} in BBH~\cite{challenging_bigbench_tasks}, Winogrande, and Hellaswag. Figure \ref{fig:orp_curve} illustrates the ORP curves before and after ILR, corresponding to either individual random factors or the considering them all. We see from the figure that ILR reduces the ORP for all settings, meaning that the evaluation is less sensitive to the change of random factors after ILR.

We attribute this observation to the following reasons. For one experiment, suppose the accuracy of an instance $x_k$ is composed of three parts:
\begin{equation}
\small
    O_M(x_k) = S_M(x_k) + \Delta S_M(x_k,f) + \epsilon_k.
\end{equation}
Here $O_M(x_k)$ is the observed accuracy of model $M$ on instance $x_k$, $S_M(x_k)$ is the inherent true accuracy of the model on this instance, $\Delta S_M(x_k, f)$ is the fluctuation term imposed by the random factors $f$, and $\epsilon_k$ is zero-mean random noise. $\Delta S_M(x_k, f)$ is related to both model $M$ and the setting of random factors $f$, which is also the source of unfair evaluations since different models may have different preferences for certain $f$. The performance score of a model is calculated as:
\begin{equation}
\label{eq:dataset_acc_with_f}
\small
    \begin{aligned}
        O_M &= \frac{1}{m}\sum_{k=1}^m \left(O_M(x_k, f) \right)\\
        &= \frac{1}{m}\sum_{k=1}^m \left( S_M(x_k) + \Delta S_M(x_k, f) + \epsilon_k \right)\\
        & = S_M + \frac{1}{m}\sum_{k=1}^m \Delta S_M(x_k, f) + \frac{1}{m}\sum_{k=1}^m\epsilon_k.
    \end{aligned}
\end{equation}
Here $S_M$ is the true accuracy of model $M$ on the dataset, $\sum_{k=1}^m\epsilon_k$ will gradually converge to $0$ as $m$ increases. However, a fixed $f$ will make $\Delta S_M(x_k, f)$ constantly positive or negative, introducing bias of evaluations and unfair comparisons between different LLMs. Fortunately, applying ILR here changes this term to $\sum_{k=1}^m \Delta S_M(x_k, f^k)$, where the effects of different $f^k$ can be either positive or negative. As $m$ increases, these effects tend to offset each other, causing this term to approach $0$ and avoiding the model bias introduced by fixing $f$. Therefore, ILR not only enhances the stability of the evaluation, but also renders each experiment less biased and fairer between models.

\subsection{ORP as a Meta-Evaluation Metric}
In addition to serving as a meta-evaluation metric for algorithms themselves, ORP can also be used as a meta-evaluation metric for datasets. A horizontal comparison of the ORP values in Figure \ref{fig:orp_curve} allows us to identify which factors have a greater impact on the dataset, thus guiding the design of algorithms to enhance robustness against them. Vertically, by comparing ORP values across different datasets under the same random factor, we can assess their robustness to random perturbations and select more robust datasets for model evaluation. Overall, ORP is a valuable meta-evaluation metric that offers deeper insights for the evaluation community. We leave further exploration of these analyses as future work.

\section{Conclusion}
This paper tackles the instability of LLM evaluations, where random factors cause score fluctuations and unreliable model rankings. We propose Instance-Level Randomization (ILR), a method that randomizes these factors for each instance and averages results across multiple experiments.

Theoretical analysis and empirical results show ILR effectively reduces variance and unfair comparisons with less than half the computational cost of previous methods. ILR lowers both instance-level and experiment-level correlations and accelerates variance reduction as the number of experiments increases. We also introduce Observed Reversal Probability (ORP) to measure ranking stability, demonstrating that ILR significantly reduces ORP, leading to more dependable evaluations.

To sum up, ILR offers a practical and efficient approach to achieve more robust, fair, and reliable LLM evaluations.

\section*{Limitations}

While the proposed Instance-Level Randomization (ILR) method demonstrates significant advantages, there are some limitations and avenues for future exploration.

First, we treat the impact of different random factors as independent ones to simplify the empirical validation of the theoretical analysis. However, literature points out that they may correlate with each other~\cite{mind_the_instructions}, causing significant difficulties to jointly analyze them. We leave the empirical validation under this circumstance in future work.

Second, the empirical validation in this study primarily focuses on a specific set of common random factors, namely few-shot examples, task descriptions, prompt formats, and option labels. Although ILR is designed to be broadly applicable to as many random factors as can be considered, its performance characteristics when applied to other, potentially more intricate or less common, random factors could be a subject for further investigation.

\clearpage
\bibliography{custom}

\clearpage
\appendix

\section{Computational Experiments}
We conduct experiments on Qwen2-72B, Llama3-70B, GLM4-9B, Llama3-8B, Gemma2-9B, Qwen2.5-7B, Qwen1.5-14B. For models with less than 14B parameters, we deploy them on 1 single A100-80GB GPU. It takes about 1 hour to run $10,000$ instances with vLLM framework. For models more than 14B parameters, we deploy them on 4 A100-80GB GPUs. It takes about 2 hours to run $10,000$ instances with vLLM framework.

\section{Information About Use Of AI Assistants}
We use GPT-4o to polish some sentences of our paper.

\section{Box plots of different LLM scores}
\label{appendix:box_plot}

\begin{figure}[htbp]
    \includegraphics[width=0.49\textwidth]{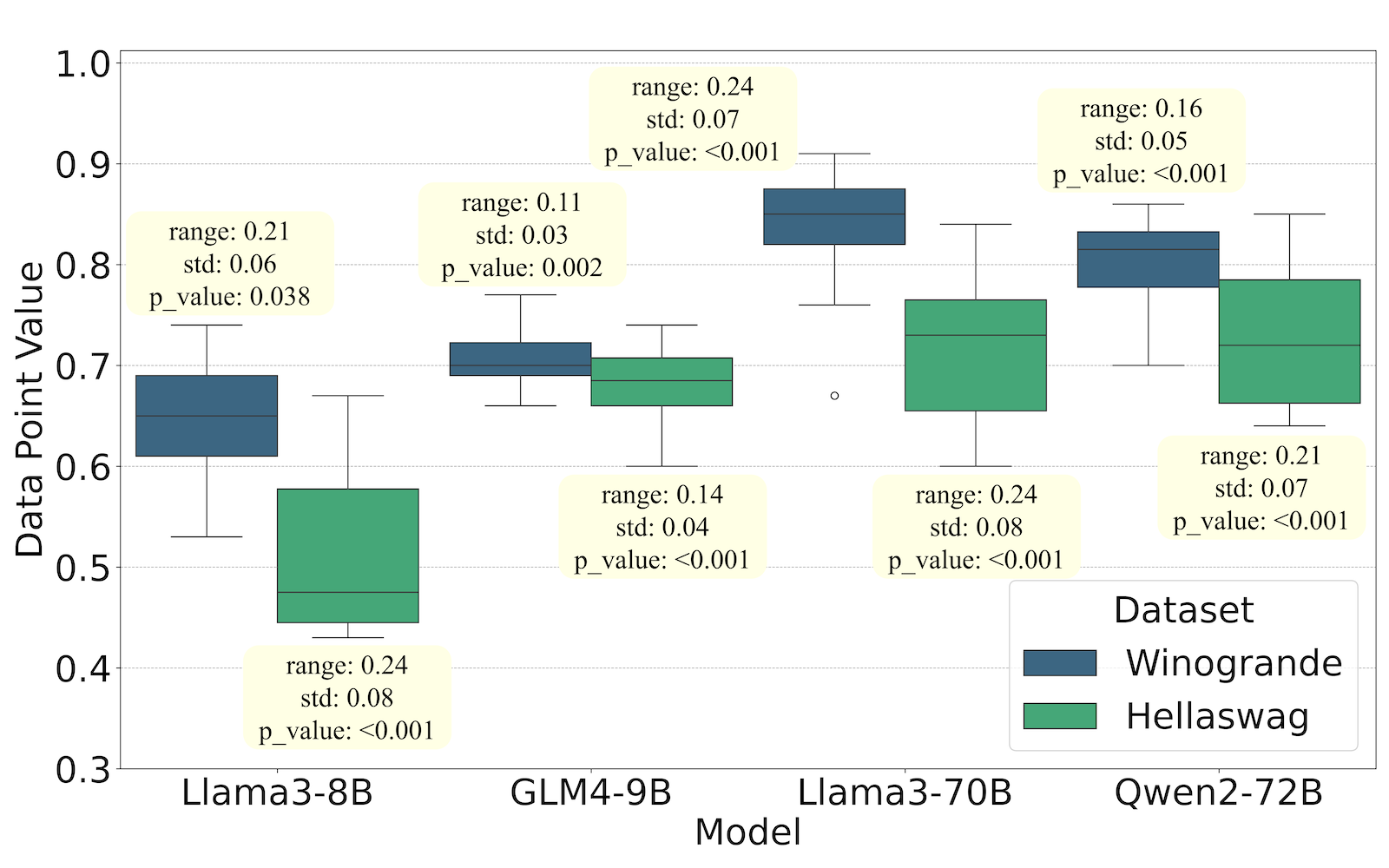}
    \centering
    \caption{Box plots of different LLMs' scores on the Winogrande and Hellaswag datasets, where each model is evaluated $8$ times under different option label settings.} 
    \label{fig:box_plot_option_labels}
\end{figure}

\begin{figure}[htbp]
    \includegraphics[width=0.49\textwidth]{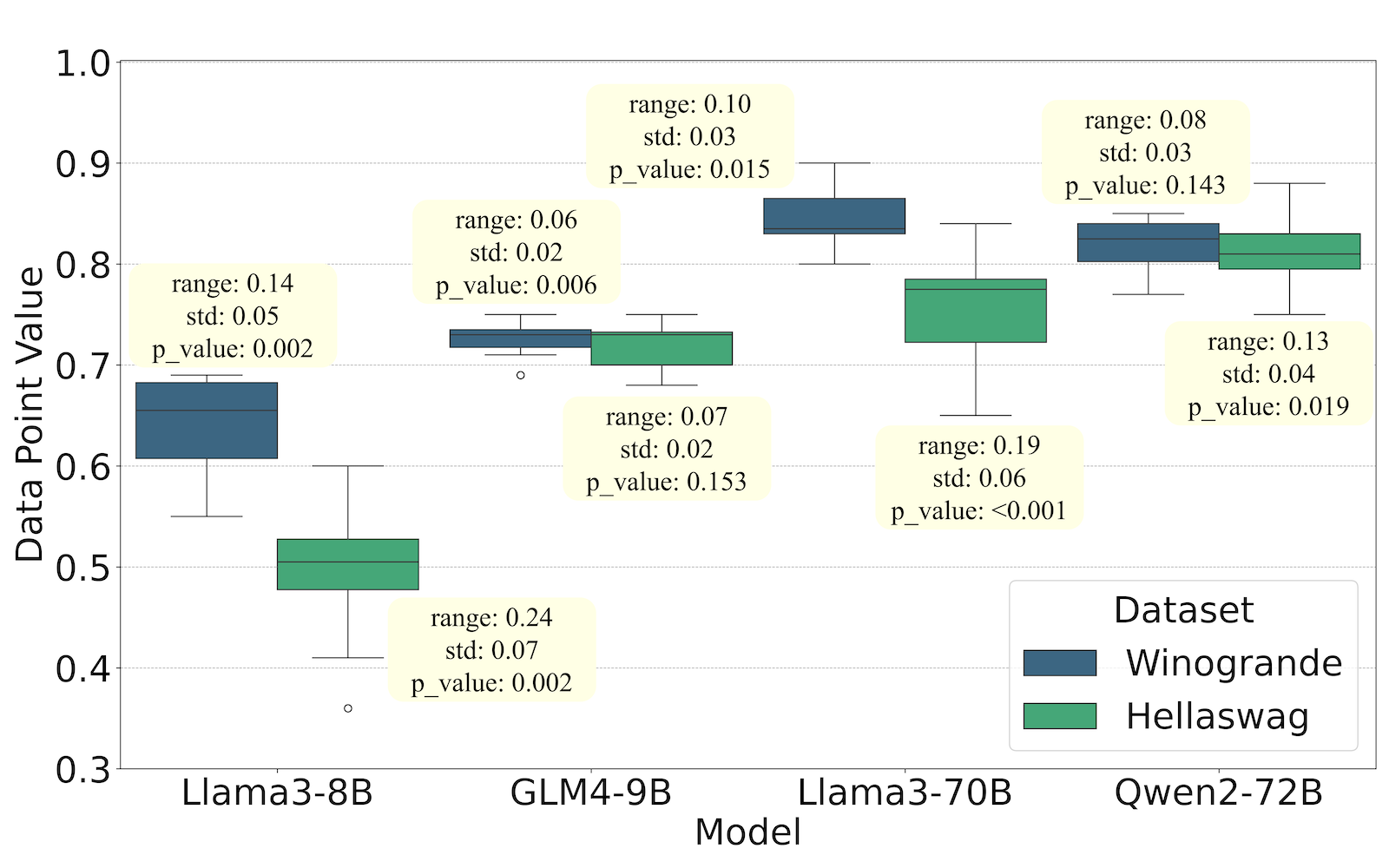}
    \centering
    \caption{Box plots of different LLMs' scores on the Winogrande and Hellaswag datasets, where each model is evaluated $8$ times under different task description settings.} 
    \label{fig:box_plot_task_descriptions}
\end{figure}

\newpage
\section{Correlations of different LLMs}
\label{appendix:corr}

\begin{figure}[htbp]
    \centering
    \subfloat{\includegraphics[width=0.32\textwidth]{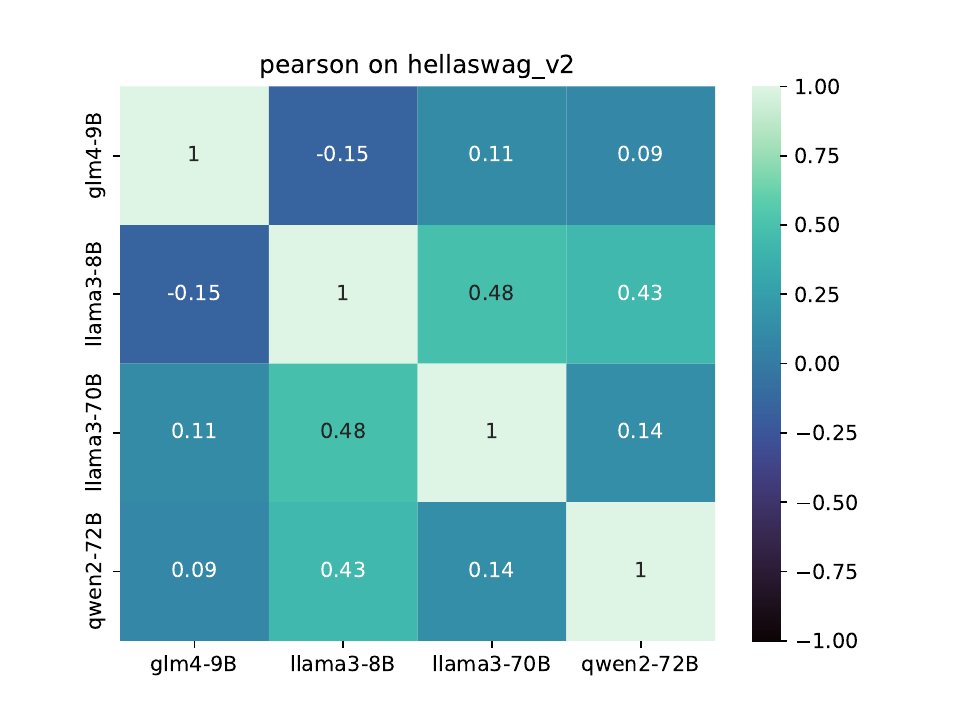}}%
    \hspace{0.01\textwidth}%
    \subfloat{\includegraphics[width=0.32\textwidth]{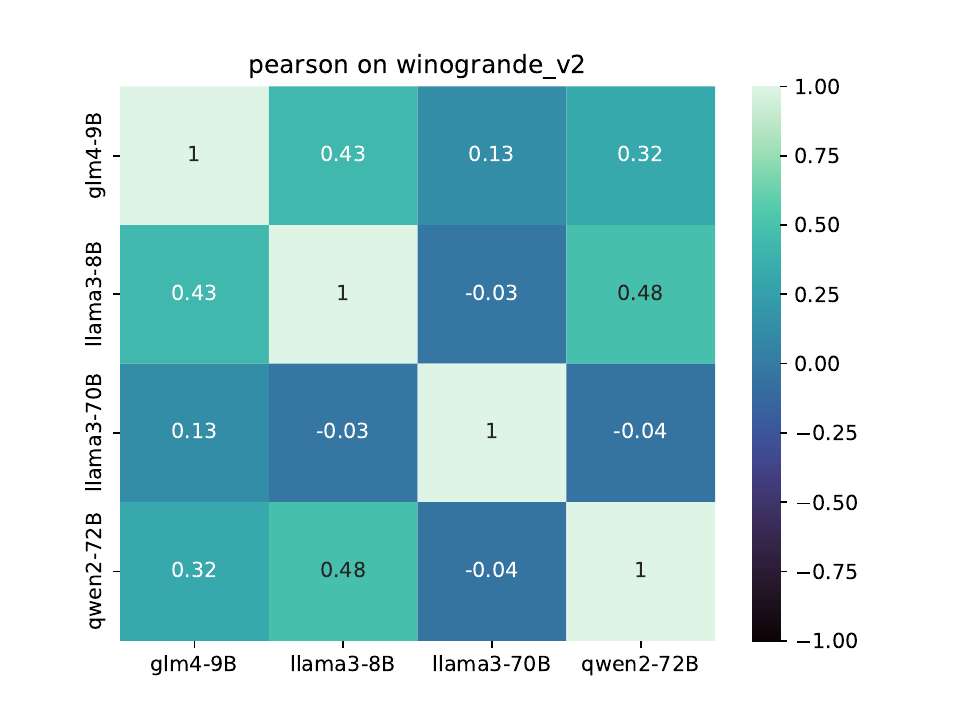}}
    \caption{Correlations of different LLMs over different option labels on Hellaswag and Winogrande.}
    \label{fig:corr_option_labels}
\end{figure}

\begin{figure}[htbp]
    \centering
    \subfloat{\includegraphics[width=0.32\textwidth]{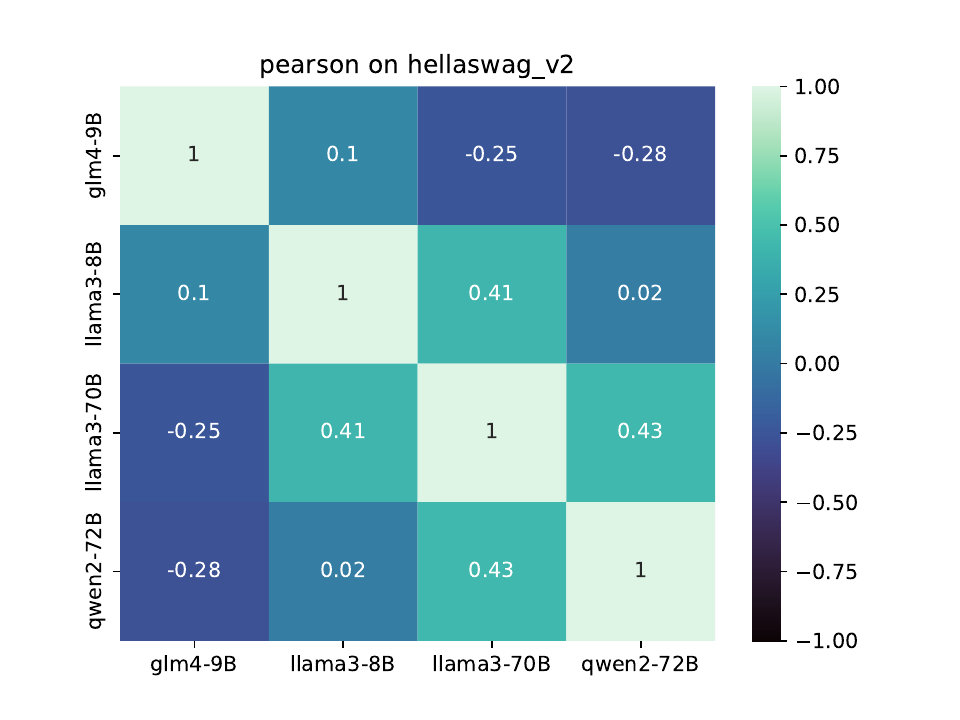}}%
    \hspace{0.01\textwidth}%
    \subfloat{\includegraphics[width=0.32\textwidth]{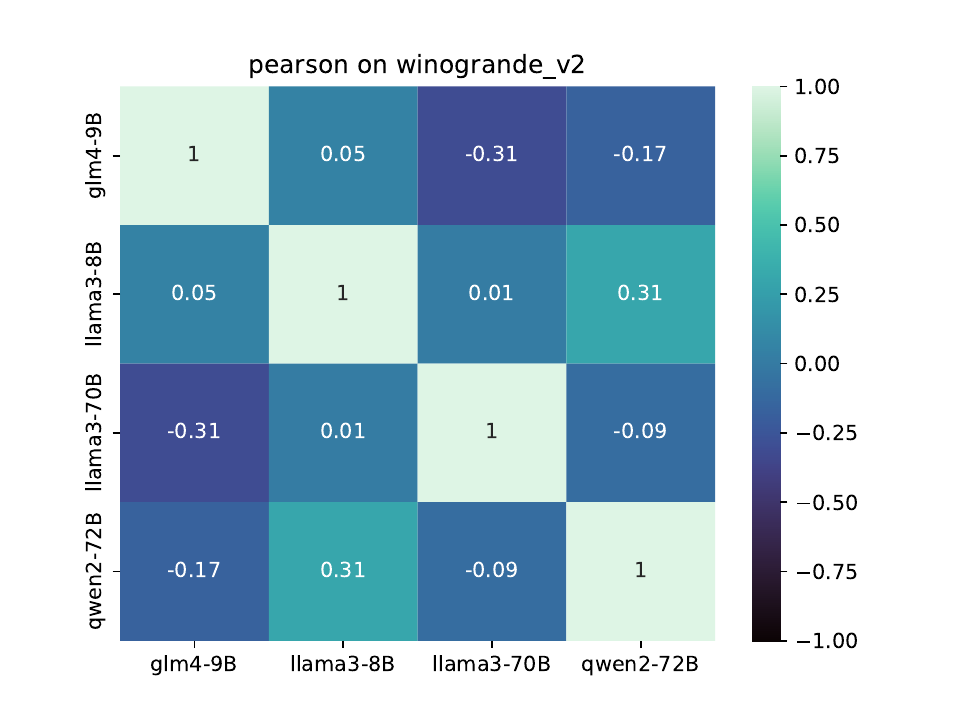}}
    \caption{Correlations of different LLMs over different task descriptions on Hellaswag and Winogrande.}
    \label{fig:corr_task_descriptions}
\end{figure}

\newpage
\section{Example of Different Random Factors}
\label{appendix:diff_random_factors}
We study few-shot examples, task descriptions, prompt formats, and option labels as the random factors that affect evaluations. As shown in Figure \ref{fig:raw_data_example}, few-shot examples and option labels are illustrated in red and green, respectively. We define the combination of question prefixes, option prefixes, and answer prefixes, as the prompt format, which is shown in blue. The combination of introductions of the task and the Chain-of-Thought~\cite{Chain-of-Thought} prompts marked in orange is defined as task descriptions.

\begin{figure}[htbp]
    \centering
\begin{tcolorbox}[title = {Example of raw data}]

\textcolor{orange}{Given a context and multiple options, choose the most reasonable continuation.}

{\color{red}Question:

[header] How to take care of a budgie $\dots$

Options:

A. Cockatiels only need a cage $\dots$

B. [substeps] They should also $\dots$

C. Cockatiels will be more interested $\dots$

D. The cage should be at least $\dots$

Let us do this task step by step.

The solution is: The question asks $\dots$

[OTHER FEW-SHOT EXAMPLES]}

\textcolor{blue}{Question:}

Several replays of acrobatic moves $\dots$

\textcolor{blue}{Options:}

\textcolor{teal}{A.} go across the screen directing $\dots$

\textcolor{teal}{B.} are written in yellow letters $\dots$

\textcolor{teal}{C.} are shown followed on the $\dots$

\textcolor{teal}{D.} are then repeated on the $\dots$

\textcolor{orange}{Let us do this task step by step.}

\textcolor{blue}{The solution is:}
\end{tcolorbox}
    \caption{The example of raw data is shown above, where the red texts are few-shot examples, the blue texts are prompt formats, the green texts are option labels, and the orange texts are task descriptions.}
    \label{fig:raw_data_example}
\end{figure}

\begin{figure}[h]
    \centering
\begin{tcolorbox}[title = {Example of different few-shot examples}]

Given a context and multiple options, choose the most reasonable continuation.

{\color{red}Question:

A soccer player is held back from $\dots$

Options:

A. are curling on the sidelines. $\dots$

B. kick the ball to one another, $\dots$

C. play soccer in an outdoor field $\dots$

D. huddle up holding a goal $\dots$

Let us do this task step by step.

The solution is: To solve this problem $\dots$

[OTHER FEW-SHOT EXAMPLES]}

Question:

Several replays of acrobatic moves $\dots$

Options:

A. go across the screen directing $\dots$

B. are written in yellow letters $\dots$

C. are shown followed on the $\dots$

D. are then repeated on the $\dots$

Let us do this task step by step.

The solution is:
\end{tcolorbox}
    \caption{The example of different few-shot examples.}
    \label{fig:different_few_shot_example}
\end{figure}

\begin{figure}[h]
    \centering
\begin{tcolorbox}[title = {Example of different prompt formats}]

Given a context and multiple options, choose the most reasonable continuation.

\textcolor{blue}{Here is a question:}

[header] How to take care of a budgie $\dots$

Options:

A. Cockatiels only need a cage $\dots$

B. [substeps] They should also $\dots$

C. Cockatiels will be more interested $\dots$

D. The cage should be at least $\dots$

Let us do this task step by step.

The solution is: The question asks $\dots$

[OTHER FEW-SHOT EXAMPLES]

\textcolor{blue}{Here is a question:}

Several replays of acrobatic moves $\dots$

\textcolor{blue}{Here are the options:}

A. go across the screen directing $\dots$

B. are written in yellow letters $\dots$

C. are shown followed on the $\dots$

D. are then repeated on the $\dots$

Let us do this task step by step.

\textcolor{blue}{Here is the solution:}
\end{tcolorbox}
    \caption{The example of different prompt formats.}
    \label{fig:different_prompt_formats}
\end{figure}

\begin{figure}[h]
    \centering
\begin{tcolorbox}[title = {Example of different option labels}]

Given a context and multiple options, choose the most reasonable continuation.

Question:

[header] How to take care of a budgie $\dots$

Options:

\textcolor{teal}{(1)} Cockatiels only need a cage $\dots$

\textcolor{teal}{(2)} [substeps] They should also $\dots$

\textcolor{teal}{(3)} Cockatiels will be more interested $\dots$

\textcolor{teal}{(4)} The cage should be at least $\dots$

Let us do this task step by step.

The solution is: The question asks $\dots$

[OTHER FEW-SHOT EXAMPLES]

Question:

Several replays of acrobatic moves $\dots$

Options:

\textcolor{teal}{(1)} go across the screen directing $\dots$

\textcolor{teal}{(2)} are written in yellow letters $\dots$

\textcolor{teal}{(3)} are shown followed on the $\dots$

\textcolor{teal}{(4)} are then repeated on the $\dots$

Let us do this task step by step.

The solution is:
\end{tcolorbox}
    \caption{The example of different option labels.}
    \label{fig:different_option_labels}
\end{figure}

\begin{figure}[h]
    \centering
\begin{tcolorbox}[title = {Example of different task descriptions}]

\textcolor{brown}{Given a context and several options, select the most logical continuation.}

Question:

[header] How to take care of a budgie $\dots$

Options:

A. Cockatiels only need a cage $\dots$

B. [substeps] They should also $\dots$

C. Cockatiels will be more interested $\dots$

D. The cage should be at least $\dots$

\textcolor{brown}{We will address this task gradually.}

The solution is: The question asks $\dots$

[OTHER FEW-SHOT EXAMPLES]

Question:

Several replays of acrobatic moves $\dots$

Options:

A. go across the screen directing $\dots$

B. are written in yellow letters $\dots$

C. are shown followed on the $\dots$

D. are then repeated on the $\dots$

\textcolor{brown}{We will address this task gradually.}

The solution is:
\end{tcolorbox}
    \caption{The example of different task descriptions.}
    \label{fig:different_task_descriptions}
\end{figure}

\end{CJK}
\end{document}